\documentclass[9pt,twocolumn,twoside]{IEEEtran}
\usepackage{spconf}
\usepackage{amsmath, amsthm, amssymb}
\usepackage{latexsym}
\usepackage{exscale}
\usepackage{fontenc}
\usepackage{graphicx}
\usepackage{epsfig}
\usepackage{nicefrac}
\usepackage{mathptmx}
\usepackage{xcolor,epic,eepic}
\usepackage[ruled,lined,linesnumbered]{algorithm2e}
\usepackage{subfigure}
\usepackage{fancyhdr}
\usepackage{hyperref}
\usepackage{multirow}
\usepackage{setspace}

\title{Statistical Compressive Sensing of Gaussian Mixture Models  \vspace{-1.5ex}}

\name{Guoshen Yu and Guillermo Sapiro \vspace{-2ex}}
\address{ECE, University of Minnesota, Minneapolis, MN 55455, U.S.A. \vspace{-2.5ex}}

\begin{document}
\renewcommand{\baselinestretch}{1.1}
\newcommand{\nin}{\in\!\!\!/}
\newcommand{\rb}{\rangle}
\newcommand{\lb}{\langle}
\newcommand{\R}{{\bf R}}
\newcommand{\thTh}{{\theta \in \Theta}}
\newcommand{\Dip}{{D}}
\newcommand{\cG}{{\cal G}}
\newcommand{\cQ}{{\cal Q}}
\newcommand{\curvelet}{{c}}
\newcommand{\orient}{{\theta}}
\newcommand{\ERC}{{\rm ERC}}
\newcommand{\DU}{{{\cal D}_U}}
\newcommand{\ipidx}{{p}}
\newcommand{\ipatom}{{g}}
\newcommand{\Ga}{{\Gamma}}
\newcommand{\pGa}{{p \in \Gamma}}
\newcommand{\lu}{{\bf l}^1}
\newcommand{\Ld}{{\bf L}^2}
\newcommand{\lz}{{\bf l}^0}
\newcommand{\ld}{{\bf l}^2}
\newcommand{\V}{{\bf V}}
\newcommand{\cL}{{\cal L}}
\newcommand{\W}{{\bf W}}
\newcommand{\La}{{\Lambda}}
\newcommand{\tLa}{{\tilde \Lambda}}
\newcommand{\pLa}{{p \in \Lambda}}
\newcommand{\qO}{{q \in \Omega}}
\newcommand{\ptLa}{{p \in \tilde \Lambda}}
\newcommand{\supp}{{\Lambda}}
\newcommand{\om}{{\omega}}
\newcommand{\mq}{{m,q}}
\newcommand{\jn}{{j,n}}
\newcommand{\opt}{{\tilde}}
\newcommand{\IPop}{{U}}
\newcommand{\CC}{{\bf C}}
\newcommand{\Z}{{\bf Z}}
\newcommand{\Reg}{\Lambda}
\newtheorem{definition}{Definition}
\newtheorem{theorem}{Theorem}
\newtheorem{lemma}{Lemma}
\newtheorem{corollary}{Corollary}
\newtheorem{prop}{Proposition}
\providecommand{\argmin}{\mathop{\textup{argmin}}}
\def\truelabel{\label}
\newcommand{\Span}{\mathop{\textup{span}}}
\newcommand{\pen}{\text{pen}}
\newcommand{\ud}{\textup{d}}
\newcommand{\D}{\mathcal{D}}
\newcommand{\Mg}{\mathcal{M}_{\gamma}}
\newcommand{\Tde}{T^{\frac{2\alpha}{\alpha+1}}}
\newcommand{\B}{\mathcal{B}}
\newcommand{\wor}{k}
\newcommand{\Cal}{\mathbf{C}^{\alpha}}
\newcommand{\eqdef}{\overset{.def}{=}}
\newcommand{\tga}{\tilde g}
\newcommand{\ldeuxj}{l^2_j}
\newcommand{\ga}{g}
\newcommand{\nogeom}{\Xi}
\newcommand{\tS}{\tilde S}
\newcommand{\tb}{\tilde b}
\newcommand{\ldeux}{l^2}
\newcommand {\ImU} {{\bf ImU}}
\newcommand{\Proba}{\mathbb{P}}
\newcommand{\C} {{\bf C}}
\newcommand{\gammageom}{\upsilon}
\newcommand{\Ch}[1]{{\bf Ch: #1}}

\newcommand{\ba}{\mathbf{a}}
\newcommand{\bbf}{\mathbf{f}}
\newcommand{\bx}{\mathbf{x}}
\newcommand{\by}{\mathbf{y}}
\newcommand{\bw}{\mathbf{w}}
\newcommand{\bz}{\mathbf{z}}

\newcommand{\bzero}{\mathbf{0}}

\newcommand{\bB}{\mathbf{B}}
\newcommand{\bD}{\mathbf{D}}
\newcommand{\bI}{\mathbf{I}}
\newcommand{\bM}{\mathbf{M}}
\newcommand{\bR}{\mathbf{R}}
\newcommand{\bS}{\mathbf{S}}
\newcommand{\bT}{\mathbf{T}}
\newcommand{\bU}{\mathbf{U}}
\newcommand{\bW}{\mathbf{W}}

\newcommand{\sample}{\Phi}
\newcommand{\dict}{\Psi}
\newcommand{\expect}{E}
\newcommand{\nsp}{\mathrm{Null}}

\maketitle

\begin{abstract}
A new framework of compressive sensing (CS), namely \textit{statistical compressive sensing} (SCS), that aims at efficiently sampling a \textit{collection} of signals that follow a statistical distribution and achieving accurate reconstruction \textit{on average}, is introduced. For signals following a Gaussian distribution, with Gaussian or Bernoulli sensing matrices of $\mathcal{O}(k)$ measurements, considerably smaller than the $\mathcal{O}(k \log(N/k))$ required by conventional CS, where $N$ is the signal dimension, and with an optimal decoder implemented with linear filtering, significantly faster than the pursuit decoders applied in conventional CS, the error of SCS is shown tightly upper bounded by a constant times the $k$-best term approximation error, with overwhelming probability. The failure probability is also significantly smaller than that of conventional CS.  Stronger yet simpler results further show that for \textit{any} sensing matrix, the error of Gaussian SCS is upper bounded by a constant times the $k$-best term approximation with probability one, and the bound constant can be efficiently calculated. For signals following Gaussian mixture models, SCS with a piecewise linear decoder is introduced and shown to produce for real images better results than conventional CS based on sparse models. 
\end{abstract}
\vspace{-0.5ex}
\begin{keywords} Compressive sensing, Gaussian mixture models  \vspace{-2ex}\end{keywords}

\section{Introduction}
\label{sec:intro}
 \vspace{-0.5ex} 
Compressive sensing (CS) attempts to achieve accurate signal reconstruction while sampling signals at a low sampling rate, typically far smaller than the Nyquist/Shannon rate. Let $\bx \in \mathbb{R}^N$ be a signal of interest, $\Phi \in \mathbb{R}^{M \times N}$ a \textit{non-adaptive} sensing matrix (\textit{encoder}), consisting of $M \ll N$ measurements, $\by = \Phi \bx \in \mathbb{R}^M$ a measured signal, and $\Delta$ a \textit{decoder} used to reconstruct $\bx$ from $\Phi \bx$. CS develops encoder-decoder pairs $(\Phi, \Delta)$ such that a small reconstruction error $\bx - \Delta(\Phi \bx)$ can be achieved. 

Reconstructing $\bx$ from $\Phi \bx$ is an ill-posed problem whose solution requires some prior information (model) on the signal. Instead of the frequency band-limit signal model assumed in classic Shannon sampling theory, conventional CS adopts a sparse signal model (manifold models have been considered as well~\cite{baraniuk2009random,chen2010compressive}), i.e., there exists a dictionary, typically an orthogonal basis $\dict \in \mathbb{R}^{N \times N}$, a linear combination of whose columns generates an accurate approximation of the signal, $\bx \approx \dict \ba$, with the coefficients $\ba[m]$, $1 \leq m \leq N$, having their amplitude decaying fast after being sorted. For signals following the sparse model, it has been shown that using some random sensing matrices such as Gaussian and Bernoulli matrices $\Phi$ with $M=\mathcal{O}(k \log(N/k))$ measurements, and an $l_1$ minimization or a greedy matching pursuit decoder $\Delta$ promoting sparsity, with high probability CS leads to accurate signal reconstruction. The obtained approximation error is tightly upper bounded by a constant times the $k$-best term approximation error, the minimum error that one may achieve by keeping the $k$ largest coefficients in $\ba$~\cite{candes2006robust,cohen2009compressed,donoho2006compressed}. 

The present paper introduces a novel framework of CS, namely \textit{statistical compressive sensing} (SCS). As opposed to conventional CS that deals
with one signal at a time, SCS aims at efficiently sampling a collection of signals and having accurate reconstruction on average. Instead of restricting to sparse models, SCS works with general Bayesian models. Assuming that the signals $\bx$ follow a distribution with probability density function (pdf) $p(\bx)$, SCS designs encoder-decoder pairs $(\sample, \Delta)$ so that the average error 
\vspace{-1ex}
\begin{equation}
\label{eqn:mean:error:scs}
 \expect_{\bx} \|\bx - \Delta(\Phi \bx) \|_X = \int  \|\bx - \Delta(\Phi \bx) \|_X p(\bx) d\bx,
\vspace{-1ex}
\end{equation}
where $ \|\cdot\|_X$ is a norm, is small. As an important example, SCS with Gaussian models is here shown to have improved performance (bounds) relative to conventional CS, the signal construction calculated with an optimal decoder $\Delta$ implemented via a fast linear filtering. Moreover, for Gaussian mixture models (GMM), SCS with a piecewise linear decoder is introduced and shown to be very effective. 

The motivation of SCS with Gaussian models is twofold. First, controlling the average error over a collection of signals is 
useful in signal acquisition, not only because one is often interested in acquiring multiple signals in real applications, but also because
 more effective processing of an individual signal, an image or a sound for example, is usually achieved by dividing the signal
 in local subparts, patches or short-time windows for instance, and a signal can be regarded as a collection 
 of subpart signals~\cite{yu2008audio, yu2010PLE}. In addition, Gaussian mixture models (GMM), which model
 signals or subpart signals with a collection of Gaussians, have been shown effective in describing real signals,
leading to excellent results in image inverse problems~\cite{yu2010PLE} and missing data estimation~\cite{leger2010Matrix}. 

After reviewing the optimal decoder for Gaussian signals in Section~\ref{sec:MAP:Gaussian}, a quick numerical check of the good performance of Gaussian SCS is first given in Section~\ref{sec:SCS:numeric:analysis}. Section~\ref{sec:SCS:analysis} analyzes this performance following a similar 
mathematical approach as the one adopted in conventional CS performance analysis~\cite{cohen2009compressed}. This result shows that 
with the same random matrices as in conventional CS, but with a considerably reduced number $M=\mathcal{O}(k)$ of measurements, and with the optimal decoder implemented with linear filtering, significantly faster than the decoders applied in conventional CS, the average error of Gaussian SCS is tightly upper bounded by a constant times the $k$-best term approximation error with overwhelming probability, the failure probability being orders of magnitude smaller than that of conventional CS. Section~\ref{sec:RIP:expect} further shows stronger yet simpler results: For \textit{any} sensing matrix, the average error of Gaussian SCS is upper bounded by a constant times the $k$-best term approximation with probability one, and the bound constant can be efficiently calculated. Section~\ref{sec:numerics} introduces, for SCS with GMM, a piecewise decoding scheme based on an efficient maximum a posteriori expectation-maximization (MAP-EM) algorithm following~\cite{yu2010PLE}, and shows better results on real images than those obtained with conventional CS. 
 
In the rest of this paper, we will assume without loss of generality Gaussian signal models $\bx \sim \mathcal{N}(\bzero, \bS)$, with mean zero and diagonal covariance matrix  $\bS = diag[\lambda_1, \ldots, \lambda_N]$, with $\lambda_1 \geq \lambda_2 \geq \ldots \geq \lambda_N$ the sorted eigenvalues. We can always center the data with respect to the Gaussian distribution and make a basis change with principal component analysis  (PCA)~\cite{yu2010PLE}. For Gaussian and Bernoulli matrices that are known to be universal, analyzing CS in the canonical basis or the PCA basis is equivalent~\cite{baraniuk2008simple}. The Gaussian distributions are assumed to be full rank, i.e., $\lambda_N > 0$, since a degenerated Gaussian can be regarded as a full-rank Gaussian within a reduced dimension. 

 \vspace{-2ex} 
\section{Optimal Decoder for Gaussian SCS}
\label{sec:MAP:Gaussian}
 \vspace{-1ex} 
It is well-known that the optimal decoders for Gaussian signals are calculated with linear filtering:
\vspace{-1.5ex}
\begin{theorem}\hspace{-1ex}~\cite{kay1998fundamentals}
\label{theo:gaussian:MAP}
Let $\bx \in \mathbb{R}^N$ be a random vector with prior pdf $\mathcal{N}(\bzero, \bS)$, and $\Phi \in \mathbb{R}^{M \times N}$, $M \leq N$, be a sensing matrix. From the measured signal $\by = \sample \bx \in \mathbb{R}^M$, the optimal decoder $\Delta$ that minimizes the mean square error (MSE)
$
 \expect_{\bx}  [\|\bx - \Delta (\Phi \bx)\|_2^2]  =  \min_{f} \expect_{\bx} [\|\bx - f(\Phi \bx)\|_2^2], \nonumber 
$
as well as the mean absolute error (MAE)
$
 \expect_{\bx}  [\|\bx - \Delta (\Phi \bx)\|_1]  =  \min_{f} \expect_{\bx} [\|\bx -  f(\Phi \bx)\|_1], \nonumber 
$
where $f$ is any mapping from $\mathbb{R}^{M}  \rightarrow \mathbb{R}^{N}$, is obtained with a linear MAP estimator,
\vspace{-2ex}
\begin{equation}
\label{eqn:MAP} 
\Delta (\Phi \bx) = \arg\max_{\bx} p(\bx|\by) = \bS \Phi^T (\Phi \bS \Phi^T)^{-1} (\Phi \bx),
\vspace{-1ex}
\end{equation}
and the resulting error $\eta = \bx - \Delta (\Phi \bx)$ is Gaussian with mean zero and with covariance matrix 
$
\Sigma_\eta  =  \expect_{\bx}  [\eta \eta^T]  
 =  \bS - \bS \Phi^T (\Phi \bS \Phi^T)^{-1} \Phi  \bS,
$
whose trace yields the MSE of SCS
\vspace{-1ex}
\begin{equation}
\label{eqn:MSE:GSCS}
 \expect_{\bx}  [\|\bx - \Delta (\Phi \bx)\|_2^2]  = Tr (\bS - \bS \Phi^T (\Phi \bS \Phi^T)^{-1} \Phi  \bS). 
 \vspace{-1ex}
 \end{equation}
\end{theorem}
In contrast to conventional CS, for which the $l_1$ minimization or greedy matching pursuit decoders, calculated with iterative procedures, have been shown optimal under certain conditions on $\Phi$ and the signal sparsity~\cite{candes2006robust, donoho2006compressed}, Gaussian SCS enjoys the advantage of having an optimal decoder~\eqref{eqn:MAP}  calculated via a \textit{linear} filtering for \textit{any} $\Phi$.

\vspace{-1ex}
\begin{corollary}
If a random matrix $\Phi \in \mathbb{R}^{M \times N}$ is drawn independently to sense each $\bx$, with all the other conditions as in Theorem~\ref{theo:gaussian:MAP}, the MSE of SCS is
\vspace{-1ex}
\begin{equation}
\label{eqn:MSE:GSCS:randPhi}
 \expect_{\bx, \Phi} \|\bx - \Delta(\Phi \bx) \|_2^2 =  \expect_{\Phi} 
[Tr (\bS - \bS \Phi^T (\Phi \bS \Phi^T)^{-1} \Phi \bS)].\vspace{-1ex}
 \end{equation}
\end{corollary}
The MSE of Gaussian SCS has closed-forms~\eqref{eqn:MSE:GSCS},~\eqref{eqn:MSE:GSCS:randPhi}, however, a mathematical analysis of~\eqref{eqn:MSE:GSCS} and~\eqref{eqn:MSE:GSCS:randPhi} seems uneasy due to the complexity of the involved matrix expression. The next section evaluates these values through Monte Carlo simulations, preceding the theoretical bounds later developed.

 \vspace{-2ex} 
\section{Performance of Gaussian SCS \\ -- a numerical analysis at first}
\label{sec:SCS:numeric:analysis}
 \vspace{-0.5ex} 
This section numerically evaluates the MSE of Gaussian SCS, and compares it with the minimal MSE generated by the best $k$-term approximation. Under the Gaussian signal model $\bx \sim \mathcal{N}(\bzero, \bS)$ with sorted eigenvalues $\lambda_1 \geq \lambda_2 \geq \ldots \geq \lambda_N$, it is well-known that the best $k$-term approximation error,
\vspace{-1ex}
\begin{equation}
\label{eqn:linear:best:k}
\sigma_{k}(\{\bx\})_X = \expect_{\bx} \|\bx - {\bx}_K \|_X~~\textrm{and}~~\sigma_{k}(\{\bx\})_2^2 = \expect_{\bx} \|\bx - {\bx}_K \|_2^2,
\vspace{-1ex}
 \end{equation}
is obtained with a \textit{linear} projection to the first $k$ entries, i.e., ${\bx}_K[n] = \bx[n]$, $\forall n \in K =\{1, \ldots, k\}$, and ${\bx}_K[n] = 0$ otherwise~\cite{mallat2008wts}. Note that $\sigma_{k}(\{\bx\})_2^2 = \sum_{m=k+1}^N \lambda_m$. This best $k$-term approximation sensing is impractical with GMM, since the Gaussian assignment of a signal is unknown at the acquisition moment, GMM describing real data much better than a single Gaussian model~\cite{yu2010PLE}. 

A power decay of the eigenvalues~\cite{mallat2008wts},
\vspace{-1ex}
\begin{equation}
\label{eqn:eigvalue:power:decay}
\lambda_m = m^{-\alpha},~~~1 \leq m \leq N,
\vspace{-1.5ex}
\end{equation}
$N=64$, is assumed in the Monte Carlo simulation. An independent random Gaussian matrix realization $\Phi$ is applied to sense each signal $\bx$~\cite{cohen2009compressed}, and~\eqref{eqn:MSE:GSCS:randPhi} is used to calculate the MSE of SCS. 

\begin{figure}[htbp]
\vspace{-2.5ex}
\begin{center}
\begin{tabular}{ccc}
\hspace{-5ex} \includegraphics[width=3.4cm]{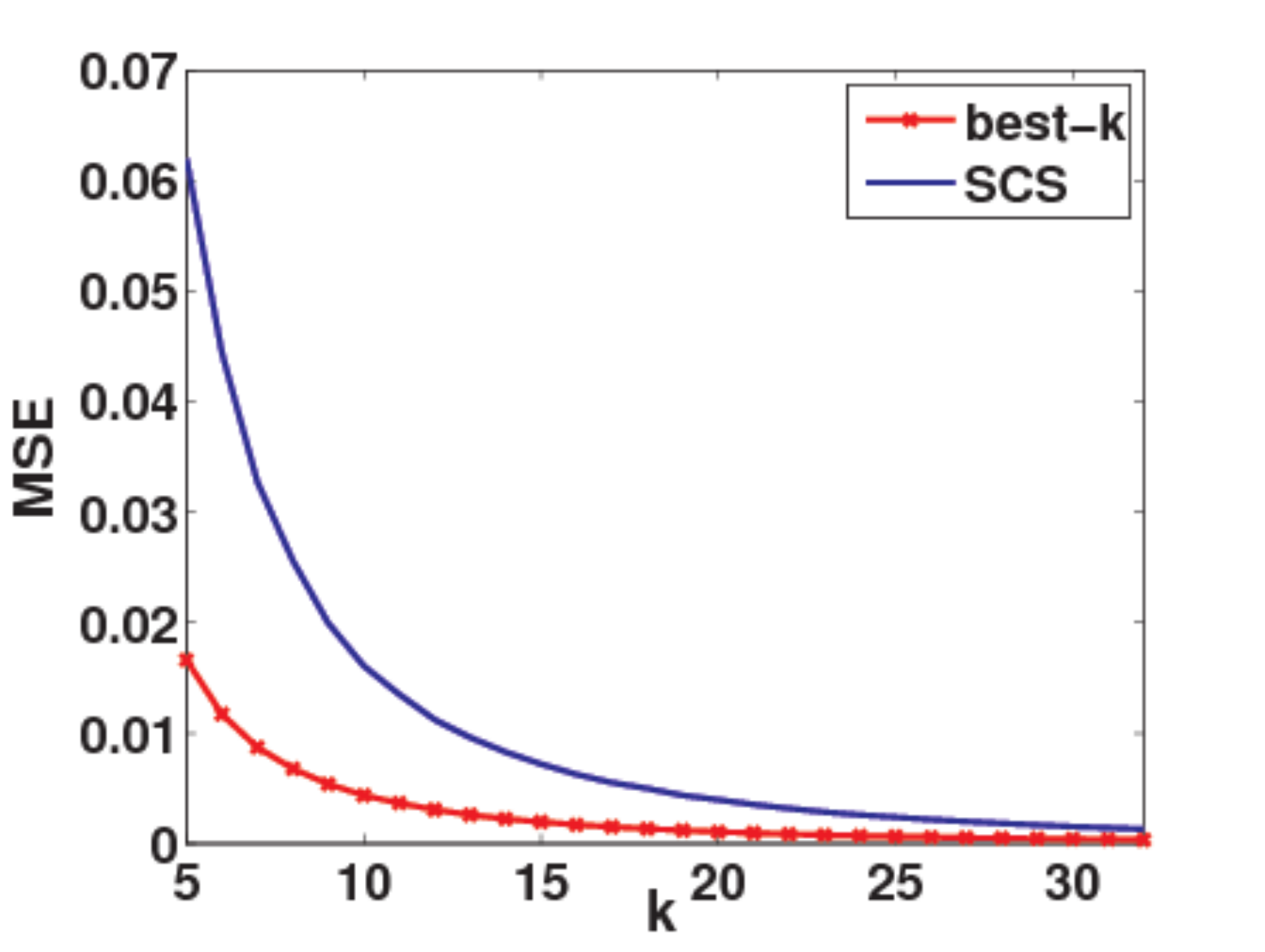} &
\hspace{-4ex} \includegraphics[width=3.4cm]{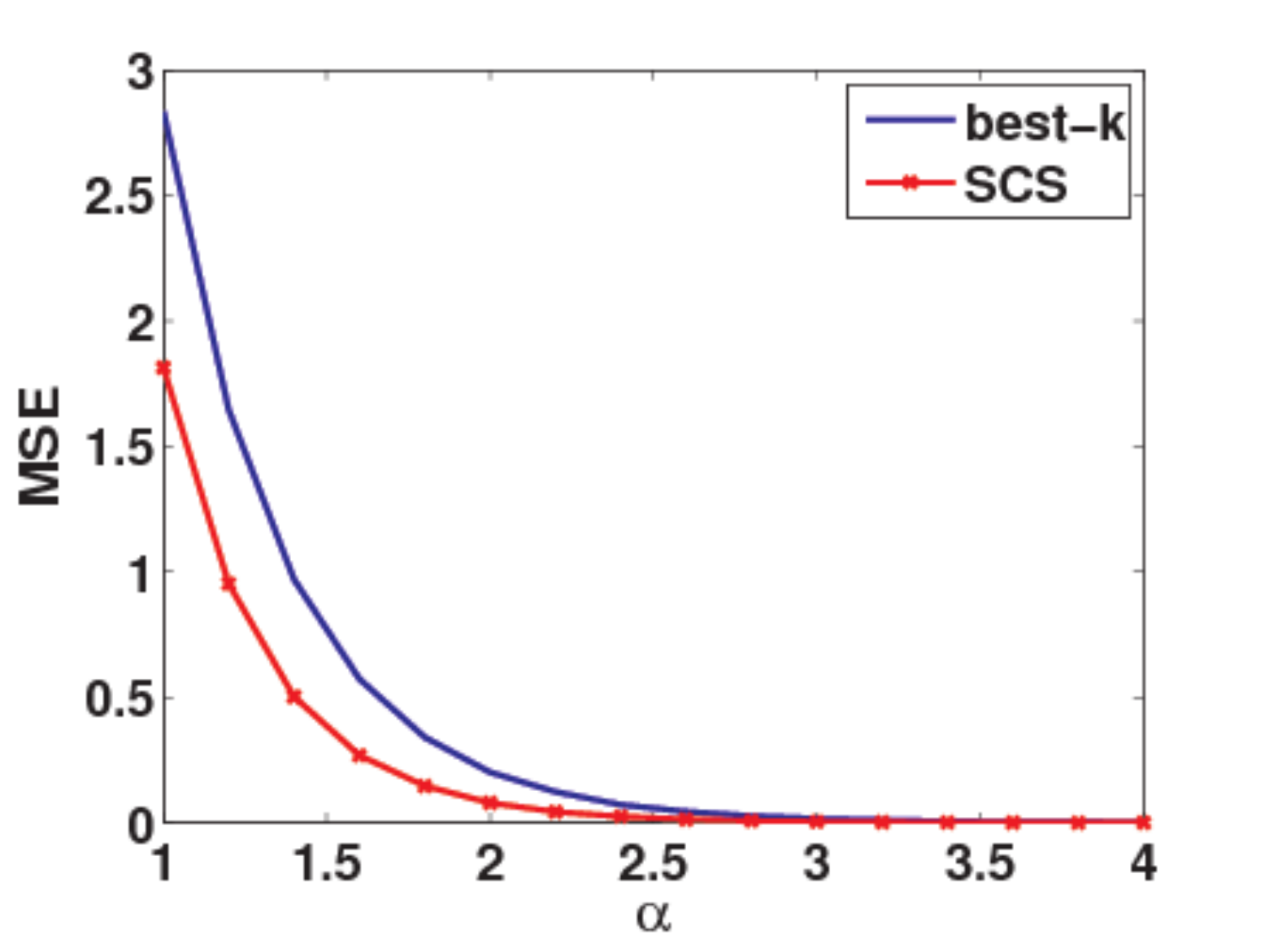} &
\hspace{-4ex} \includegraphics[width=3.4cm]{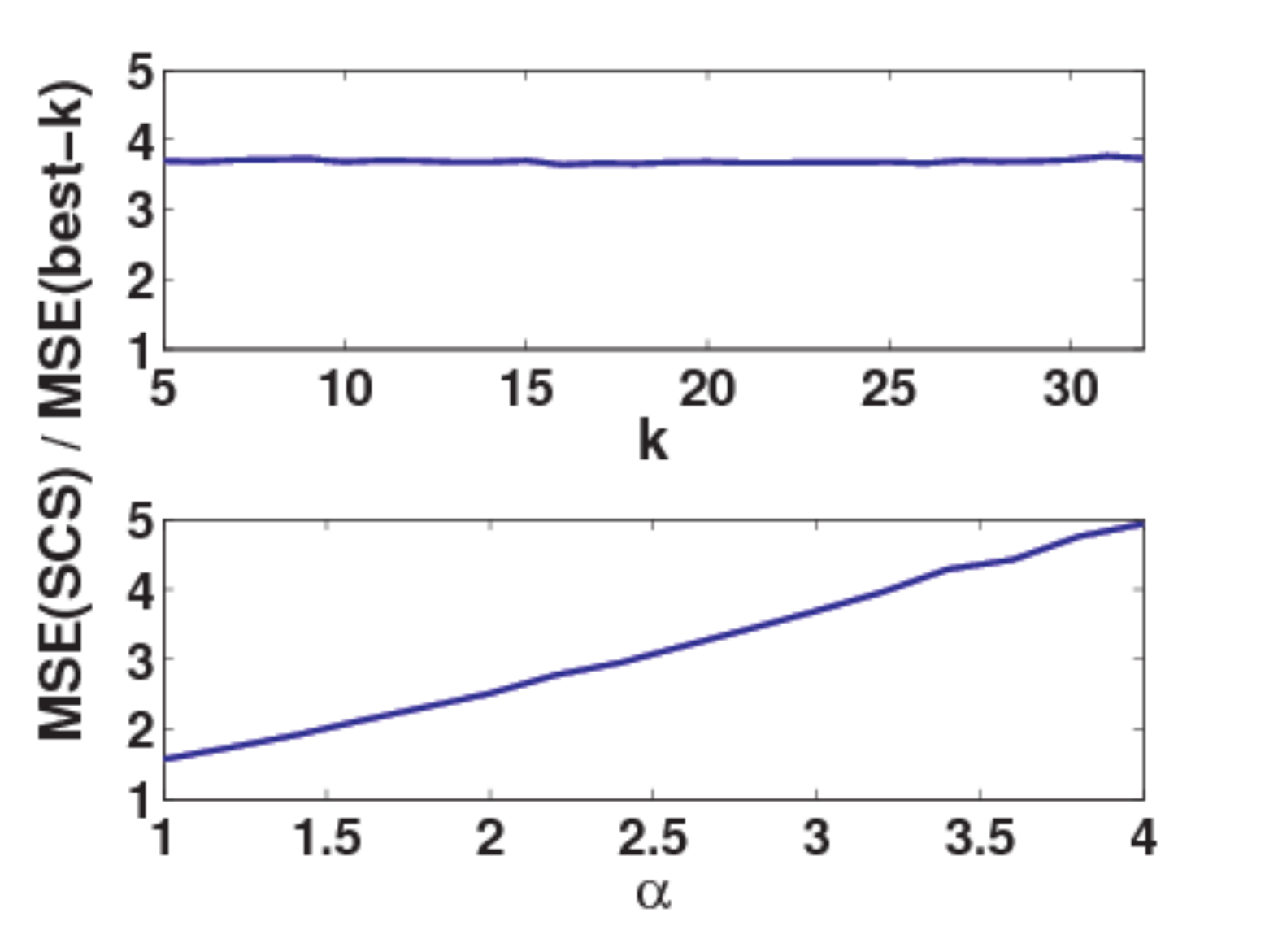} \vspace{-1ex} \\
\hspace{-3ex}\textbf{\small(a)}&  \hspace{-3ex}\textbf{\small(b)} &  \hspace{-3ex}\textbf{\small(c)}  \vspace{-0.5ex} \\
\end{tabular}
\end{center}
\vspace{-3ex}
\caption{\small Comparison of the MSE of SCS and that of the best $k$-term approximation for Gaussian signals. See text for details.} \label{fig:MSE:scs:vs:bestk}
\vspace{-2ex}
\end{figure}

Figures~\ref{fig:MSE:scs:vs:bestk} (a) and (c)-top plot the MSE of SCS and that of the best $k$-term approximation, as well as their ratio as a function of $k$ for Gaussian signals with a typical eigenvalue decay $\alpha=-3$. With $k$ increasing, both MSEs decrease, their ratio being almost constant at about $3.7$. The same is plotted in figures~\ref{fig:MSE:scs:vs:bestk} (b) and (c)-bottom, with $k$ fixed at a typical value of $10$, and $\alpha$ varying from $1$ to $4$. When $\alpha$ increases, the eigenvalues decay faster, and the MSEs for both methods decrease. The ratio increases almost linearly with $\alpha$. 

These results indicate a good performance of Gaussian SCS, its MSE is only a small number of times larger than that of the best $k$-term approximation. The next sections provide mathematical analysis of this performance. 

 \vspace{-2ex} 
\section{Performance Bounds of Gaussian SCS}
\label{sec:SCS:analysis}
 \vspace{-0.5ex} 
Following~\cite{cohen2009compressed}, this section shows that with Gaussian and Bernoulli random matrices of $\mathcal{O}(k)$ measurements, considerably smaller than the $\mathcal{O}(k \log (N/k))$ required by conventional CS, the average error of Gaussian SCS is tightly upper bounded by a constant times the $k$-best term approximation with overwhelming probability, the failure probability being orders of magnitude smaller than that of conventional CS. The proofs of the theorems will not be given due to the space limitations. We consider only the encoder-decoder pairs $(\Phi, \Delta)$ that preserve $\Phi \bx$, i.e., $\Phi (\Delta (\Phi \bx)) = \Phi \bx$, satisfied by the optimal $\Delta$ in~\eqref{eqn:MAP} for Gaussian signals $\bx$,  $\forall \Phi$. 

 \vspace{-2ex} 
\subsection{From Null Space Property to Instance Optimality}
The \textit{instance optimality in expectation} bounds the average error of SCS with a constant times that of the best $k$-term approximation~\eqref{eqn:linear:best:k}, defining the desired SCS performance:
\vspace{-1ex}
\begin{definition} 
We say that $(\Phi, \Delta)$ is \textit{instance optimal in expectation} of order $k$ in $\|\cdot\|_X$, with a constant $C_0$, if 
\vspace{-1ex}
\begin{equation}
\label{eqn:instance:optimality}
E_{\bx, (\Phi)} \|\bx - \Delta (\Phi \bx)\|_X \leq C_0 \sigma_{k}(\{\bx\})_X,
\vspace{-1ex}
\end{equation}
the expectation with respect to $\bx$, and to $\Phi$ if one random $\Phi$ is drawn independently for each $\bx$. Similarly, the MSE instance optimality of order $k$ is defined as 
\vspace{-1ex}
\begin{equation}
\label{eqn:instance:optimality:MSE}
E_{\bx, (\Phi)} \|\bx - \Delta (\Phi \bx)\|_2^2 \leq C_0 \sigma_{k}(\{\bx\})_2^2.
\vspace{-1ex}
\end{equation}
\end{definition}

The \textit{null space property in expectation} defined next will be shown equivalent to the instance optimality in expectation.
\vspace{-1ex}
\begin{definition} 
We say that $\Phi$ in $(\Phi, \Delta)$ has the null space property  in expectation of order $k$ in $\|\cdot\|_X$, with constant $C$, if 
\vspace{-1ex}
\begin{equation}
\label{eqn:null:space:property}
E_{\bx, (\Phi)} \|\eta\|_X \leq C \sigma_{k}(\{\eta\})_X,~~~\forall \eta = \bx -  \Delta (\Phi \bx),
\vspace{-1ex}
\end{equation}
the expectation with respect to $\bx$, and to $\Phi$ if one random $\Phi$ is drawn independently for each $\bx$. Note that $\eta \in \nsp(\Phi)$. Similarly, the MSE null space property of order $k$ is defined as 
\vspace{-1ex}
\begin{equation}
\label{eqn:null:space:property:MSE}
E_{\bx, (\Phi)} \|\eta\|_2^2 \leq C \sigma_{k}(\{\eta\})_2^2,~~~\forall \eta = \bx -  \Delta (\Phi \bx).
\vspace{-1ex}
\end{equation}
\end{definition}
\vspace{-1ex}
\begin{theorem}
\label{theo:nullspace2optimality}
Given an $M \times N$ matrix $\Phi$, a norm $\|\cdot\|_X$, and a positive integer $k \le N$, a sufficient condition that there exists a decoder $\Delta$ such that the instance optimality in expectation~\eqref{eqn:instance:optimality} holds with constant $C_0$, is that the null space property in expectation~\eqref{eqn:null:space:property} holds with $C=C_0/2$ for this $(\Phi, \Delta)$. A necessary condition is the null space property in expectation~\eqref{eqn:null:space:property} with $C = C_0$. Similar results hold between MSE instance optimality~\eqref{eqn:instance:optimality:MSE} and null space property~\eqref{eqn:null:space:property:MSE}, with the constant $C=C_0/4$ in the sufficient condition.
\vspace{-3ex}
\end{theorem}
Comparing to conventional CS that requires the null space property to hold with the best $2k$-term approximation error~\cite{cohen2009compressed}, the requirement for Gaussian SCS is relaxed to $k$, thanks to the linearity of the best $k$-term approximation for Gaussian signals. 

Theorem~\ref{theo:nullspace2optimality} proves the existence of the decoder $\Delta$ for which the instance optimality in expectation holds for $(\Phi, \Delta)$, given the null space property in expectation. However, it does not explain how such decoder is implemented. The following Corollary, a direct consequence of theorems~\ref{theo:gaussian:MAP} and~\ref{theo:nullspace2optimality}, shows that for Gaussian signals the optimal decoder~\eqref{eqn:MAP} leads to the instance optimality in expectation. 
\vspace{-1ex}
\begin{corollary}
\label{corollary:nullspace2optimality:gaussian} 
For Gaussian signals $\bx \sim \mathcal{N}(\bzero, \bS)$, if an $M \times N$ sensing matrix $\Phi$ satisfies the null space property in expectation~\eqref{eqn:null:space:property} of order $k$ in $\|\cdot\|_1$, with constant $C_0/2$, or the MSE null space property~\eqref{eqn:null:space:property:MSE} of order $k$ with constant $C_0/4$, then the optimal and linear decoder $\Delta = \bS \Phi^T (\Phi \bS \Phi^T)^{-1}$ satisfies the instance optimality in expectation~\eqref{eqn:instance:optimality} in $\|\cdot\|_1$, or the MSE instance optimality~\eqref{eqn:instance:optimality:MSE}. 
\end{corollary}
 \vspace{-4ex} 
\subsection{From RIP to Null Space Property}
The Restricted Isometry Property (RIP) of a matrix measures its ability to preserve distances, and is related to the null space property in conventional CS~\cite{candes2006near, donoho2006compressed}. The new \textit{linear} RIP of order $k$ restricts the requirement of conventional RIP of order $k$, to a union of $k$-dimensional linear subspaces with consecutive supports:
\vspace{-1ex}
\begin{definition}
Let $k \leq N$ be a positive integer. Let $K_1$ define a linear subspace of functions with support in the first $k$ indices in $[1,N]$, $K_2$ a linear subspace of functions with support in the next $k$ indices, and so on. The functions in the last linear subspace $K_J$ defined this way may have support with less than $k$ indices. An $M \times N$ matrix $\Phi$ is said to have linear RIP of order $k$ with constant $\delta$ if 
\vspace{-1ex}
\begin{equation}
\label{eqn:linear:RIP}
(1-\delta) \|\bx\|_2 \leq \|\Phi \bx\|_2 \leq (1+\delta) \|\bx\|_2,~~~\forall~\bx \in \cup_{j=1}^J K_j.
\vspace{-1ex}
\end{equation}
\end{definition}
The linear RIP is a special case of the block RIP~\cite{eldar2009robust},  with block sparsity one
and blocks having consecutive support of the same size.

The following theorem relates the linear RIP~\eqref{eqn:linear:RIP} of a matrix $\Phi$ to its null space property in expectation~\eqref{eqn:null:space:property}. 
\vspace{-1ex}
\begin{theorem}
\label{theo:RIP2NullSpace}
Let $\bx \in \mathbb{R}^N$ be a random vector that follows a certain distribution. Let $\Phi$ be an $M \times N$ matrix that satisfies the linear RIP of order $2k$ with $\delta < 1$, and let $\Delta$ be a decoder. Let $\eta = \bx - \Delta (\Phi \bx)$.  Assume further that $E_{\bx, (\Phi)}|\eta[n]|$ decays in $n$: $E_{\bx, (\Phi)}|\eta[n+1]| < E_{\bx, (\Phi)}|\eta[n]|$, $\forall n < N$. Then $\Phi$ satisfies the null property in expectation of order $k$ in $\|\cdot\|_1$~\eqref{eqn:null:space:property}, with constant $C_0 = 1 + k^{1/2} \frac{1+\delta}{1-\delta}$.
\vspace{-1ex}
\end{theorem}
For Gaussian signals $\bx \in \mathcal{N}(\bzero, \bS)$,  with $\Phi$ Gaussian or Bernoulli matrices, one realization drawn independently for each $\bx$, and with $\Delta$ the optimal decoder~\eqref{eqn:MAP}, the decay of  $E_{\bx, \Phi}|\eta[n]|$ assumed in 
in Theorem~\ref{theo:RIP2NullSpace} is verified through Monte Carlo simulations. 
 \vspace{-3ex} 
\subsection{From Random Matrices to RIP}
The next Theorem shows that Gaussian and Bernoulli matrices satisfy the RIP for \textit{one} subspace with overwhelming probability. 
\vspace{-1ex}
\begin{theorem}~\cite{baraniuk2008simple}
\label{theo:RIP:prob:oneset}
Let $\Phi$ be a random matrix of size $M \times N$ drawn according to any distribution that satisfies the concentration inequality
\vspace{-1ex}
\begin{equation}
\label{eqn:concentration:inequatliy}
\textrm{Pr}(\|\Phi \bx\|_2^2 - \|\bx\|_2^2| \geq \epsilon \|\bx\|_2^2) \leq 2 e^{-M c_0(\delta/2)},~~~\forall~\bx \in \mathbb{R}^N,
\vspace{-1ex}
\end{equation}
where $0 < \delta < 1$, and $c_0(\delta/2) > 0$ is a constant depending only on $\delta/2$. Then for any set $K \subset \{1,\ldots,N\}$ with $|K|=k<M$, we have
\vspace{-1ex}
\begin{equation}
\label{eqn:linear:RIP:oneset}
(1-\delta) \|\bx\|_2 \leq \|\Phi \bx\|_2 \leq (1+\delta) \|\bx\|_2,~~~\forall~\bx \in \mathcal{X}_K,
\vspace{-1ex}
\end{equation}
where $\mathcal{X}_K$ is the set of all vectors in $\mathbb{R}^N$ that are zero outside of $K$, with probability greater than or equal to
$
1 - 2(12/\delta)^k e^{-c_0(\delta/2)M}.
$
Gaussian and Bernoulli matrices satisfy the concentration inequality~\eqref{eqn:concentration:inequatliy}.
\vspace{-1ex}
\end{theorem}
The linear RIP of order $k$~\eqref{eqn:linear:RIP} requires that~\eqref{eqn:linear:RIP:oneset} holds for $N/k \le N$ subspaces. The next Theorem follows from Theorem~\ref{theo:RIP:prob:oneset} by simply multiplying by $N$ the probability that the RIP fails to hold for one subspace. 
\vspace{-1ex}
\begin{theorem}
Suppose that $M$, $N$ and $0 < \delta < 1$ are given. Let $\Phi$ be a random matrix of size $M \times N$ drawn according to any distribution that satisfies the concentration inequality~\eqref{eqn:concentration:inequatliy}. Then there exist constants $c_1, c_2 > 0$ depending only on $\delta$ such that the linear RIP of order $k$~\eqref{eqn:linear:RIP} holds  with probability greater than or equal to $1- 2 N e^{-c_2 M}$ for $\Phi$ with the prescribed $\delta$ and $k \leq c_1 M$.
\vspace{-1ex}
\end{theorem}
Comparing with conventional CS, where the null space property requires that the RIP~\eqref{eqn:linear:RIP:oneset} holds for $\binom{N}{k}$ subspaces~\cite{baraniuk2008simple, candes2006near, donoho2006compressed}, the number of subspaces in the linear RIP~\eqref{eqn:linear:RIP} is sharply reduced to $N/k$ for Gaussian SCS. In consequence, with the same number of measurements $M$, the probability that a Gaussian or Bernoulli matrix $\Phi$ satisfies the linear RIP is substantially higher than that for the conventional RIP. Equivalently, given the same probability that $\Phi$ satisfies the linear RIP or the conventional RIP of order $k$, the required number of measurements for the linear RIP is $M \sim \mathcal{O}(k)$, substantially smaller than the $M \sim \mathcal{O}(k \log(N/k))$ required for the conventional RIP. Similar improvements have been obtained with model-based CS that assumes structured sparsity on the signals~\cite{baraniuk2010model}. 

With the results above, we have showed that for Gaussian signals, with sensing matrices satisfying the linear RIP~\eqref{eqn:linear:RIP} of order $2k$, for example Gaussian or Bernoulli matrices with $\mathcal{O}(k)$ rows, with overwhelming probability, and with the optimal decoder~\eqref{eqn:MAP}, SCS leads to the instance optimality in expectation  of order $k$ in $\|\cdot\|_1$~\eqref{eqn:instance:optimality}, with constant $C_0 = 2(1 +  k^{1/2} \frac{1+\delta}{1-\delta})$. By the definition of CS, $k^{1/2}$ is typically small.
 \vspace{-2ex}   
\section{Performance Bounds of Gaussian SCS \\ with RIP in Expectation}
 \vspace{-1ex} 
\label{sec:RIP:expect}
This section shows that with an {\it RIP in expectation}, a matrix isometry property more adapted to SCS, the Gaussian SCS MSE instance optimality~\eqref{eqn:instance:optimality:MSE} of order $k$ and constant $C_0$, holds in the $l_2$ norm with probability one for {\it any} matrix. $C_0$ has a closed-form and can be easily computed numerically.
\vspace{-1ex}
\begin{definition}
\label{def:RIP:expect}
Let $\bx \in \mathbb{R}^N$ be a random vector that follows a certain distribution. Let  $\Phi$ be an $M \times N$ sensing matrix and let $\Delta$ be a decoder. Let $\eta = \bx - \Delta (\Phi \bx)$. $\Phi$ in  $(\Phi, \Delta)$ is said to have RIP in expectation in $K$ with constant $c_K$ if 
\vspace{-1ex}
\begin{equation}
\label{eqn:RIP:expect}
{E_{\bx, (\Phi)} \|\Phi \eta_K\|_2^2} = c_K {E_{\bx, (\Phi)} \|\eta_K\|_2^2},~~~\forall~\eta = \bx - \Delta (\Phi \bx),
\vspace{-1ex}
\end{equation}
where $K \subset \{1, \ldots, N\}$, $\eta_K \in \mathbb{R}^N$ is the signal $\eta$ restricted to $K$ ($\eta_K[n]=\eta[n],~\forall~n \in K$, and $0$ otherwise), and the expectation is with respect to $\bx$, and to $\Phi$ if one random $\Phi$ is drawn independently for each $\bx$.
\vspace{-1ex}
\end{definition}

The conventional RIP is known to be satisfied  only by some random matrices, Gaussian and Bernoulli matrices for example, with high probability. For a given matrix, checking the RIP property is however NP-hard~\cite{baraniuk2008simple}. By contrast, the constant of the RIP in expectation~\eqref{eqn:RIP:expect} can be measured for \textit{any} matrix via a fast simulation, the quick convergence guaranteed by the concentration of measure~\cite{talagrand1996new}. The next proposition, directly following from~\eqref{eqn:MSE:GSCS} and~\eqref{eqn:MSE:GSCS:randPhi}, further shows that for Gaussian signals, the RIP in expectation has its constant in a closed form. 
\vspace{-1ex}
\begin{prop}
\label{prop:RIP:expect:gassian}
Assume $\bx \sim \mathcal{N}(\bzero, \bS)$, $\Phi$ is an $M \times N$ sensing matrix and $\Delta$ is the optimal decoder~\eqref{eqn:MAP}. Then $\Phi$ in  $(\Phi, \Delta)$ satisfies the RIP in expectation in $K$, 
\vspace{-3ex}
{\small
\begin{equation}
\label{eqn:RIP:gauss}
  {(E_\Phi) \left[Tr\left( \Phi \bR_K \bS \bR_K^T \Phi^T - \Phi \bR_K \bS \Phi^T (\Phi \bS \Phi^T)^{-1} \Phi \bS \bR_K^T \Phi^T\right) \right]} \nonumber \vspace{-1ex}
 \end{equation}
 \begin{equation} 
   = c_K
{(E_\Phi) \left[ Tr\left(\bR_K \bS \bR_K^T - \bR_K \bS \Phi^T (\Phi \bS \Phi^T)^{-1} \Phi \bS \bR_K^T\right) \right]} \vspace{-1.5ex},
\end{equation}
}where $\bR_K$ is an $N \times N$ extraction matrix giving $\eta_K = \bR_K \eta$, i.e., $\bR_K(i,i)=1$, $\forall i \in K$, all the other entries being zero. The expectation with respect to $\Phi$ is calculated if one random $\Phi$ is drawn independently for each $\bx$. 
\vspace{-1ex}
\end{prop}

The next Theorem shows that the RIP in expectation leads to the MSE null space property holding in equality. 
\vspace{-1ex}
\begin{theorem}
\label{theo:RIP:expect:null:space}
Let $\bx \in \mathbb{R}^N$ be a random vector that follows a certain distribution,  $\Phi$ an $M \times N$ sensing matrix, and $\Delta$ a decoder. 
Assume ${E_{\bx, (\Phi)}  \|\eta_{K}\|_2^2} \neq 0$ and ${E_{\bx, (\Phi)}  \|\eta_{K^C}\|_2^2} \neq 0$, for some $K \subset \{1, \ldots, N\}$. Assume that $\Phi$ in $(\Phi, \Delta)$ has the RIP in expectation in $K$ with constant $a_K > 0$, and in $K^C =  \{1, \ldots, N\}\backslash K$  with constant $b_K > 0$:
\vspace{-1ex}
{\small
\begin{equation}
\label{eqn:RIP:expt}
\frac{E_{\bx, (\Phi)} \|\Phi \eta_{K}\|_2^2}{E_{\bx, (\Phi)}  \|\eta_{K}\|_2^2} = a_K,~~~\frac{E_{\bx, (\Phi)}  \|\Phi \eta_{K^C}\|_2^2}{E_{\bx, (\Phi)}  \|\eta_{K^C}\|_2^2} = b_K,
~~~\forall~\eta = \bx - \Delta \Phi \bx. \vspace{-1ex}
\end{equation}
}Then $\Phi$ satisfies
\vspace{-1ex}
\begin{equation}
\label{eqn:null:space:equality1}
E_{\bx, (\Phi)}\|\eta\|_2^2 = C_0 E_{\bx, (\Phi)}\|\eta_{K^C}\|_2^2,
\vspace{-1ex}
\end{equation}
where $C_0 = 1 + {b_K}/{a_K}$. In particular, if $K = \{1, \ldots, k\}$, then  $\Phi$ satisfies the MSE null space property of order $k$, which holds with equality,
\vspace{-1ex}
\begin{equation}
\label{eqn:null:space:equality}
E_{\bx, (\Phi)}\|\eta\|_2^2 = C_0 \sigma_k(\{\eta\})_2^2.
\vspace{-1ex}
\end{equation}
\end{theorem}
Let us check the MSE null space property constant $C_0$ of different sensing matrices in SCS for Gaussian signals $\bx \in \mathbb{R}^N \sim \mathcal{N}(\bzero, \bS)$, assuming the eigenvalues of $\bS$ follow a power decay~\eqref{eqn:eigvalue:power:decay} with $\alpha=-3$, and $N=64$. Gaussian, Bernoulli and random subsampling matrices $\Phi$ of size $M \times N$ are considered, and the optimal and linear decoder $\Delta$~\eqref{eqn:MAP} is applied to reconstruct the signals. For each matrix distribution, a different random matrix realization $\Phi$ is applied to sense each signal $\bx$. Note that since the random subsampling matrix $\Phi$, each row containing one entry with value 1 at a random position and 0 otherwise, has the maximal coherence with the canonical basis, this matrix is not suitable for directly sensing $\bx$~\cite{candes2007sparsity}, and is replaced by $\Phi \dict$ in the simulation, with $\dict$ a DCT basis having low coherence with $\Phi$. 

\begin{figure}[htbp]
\vspace{-1ex}
\begin{center}
\begin{tabular}{cc}
\includegraphics[width=4cm]{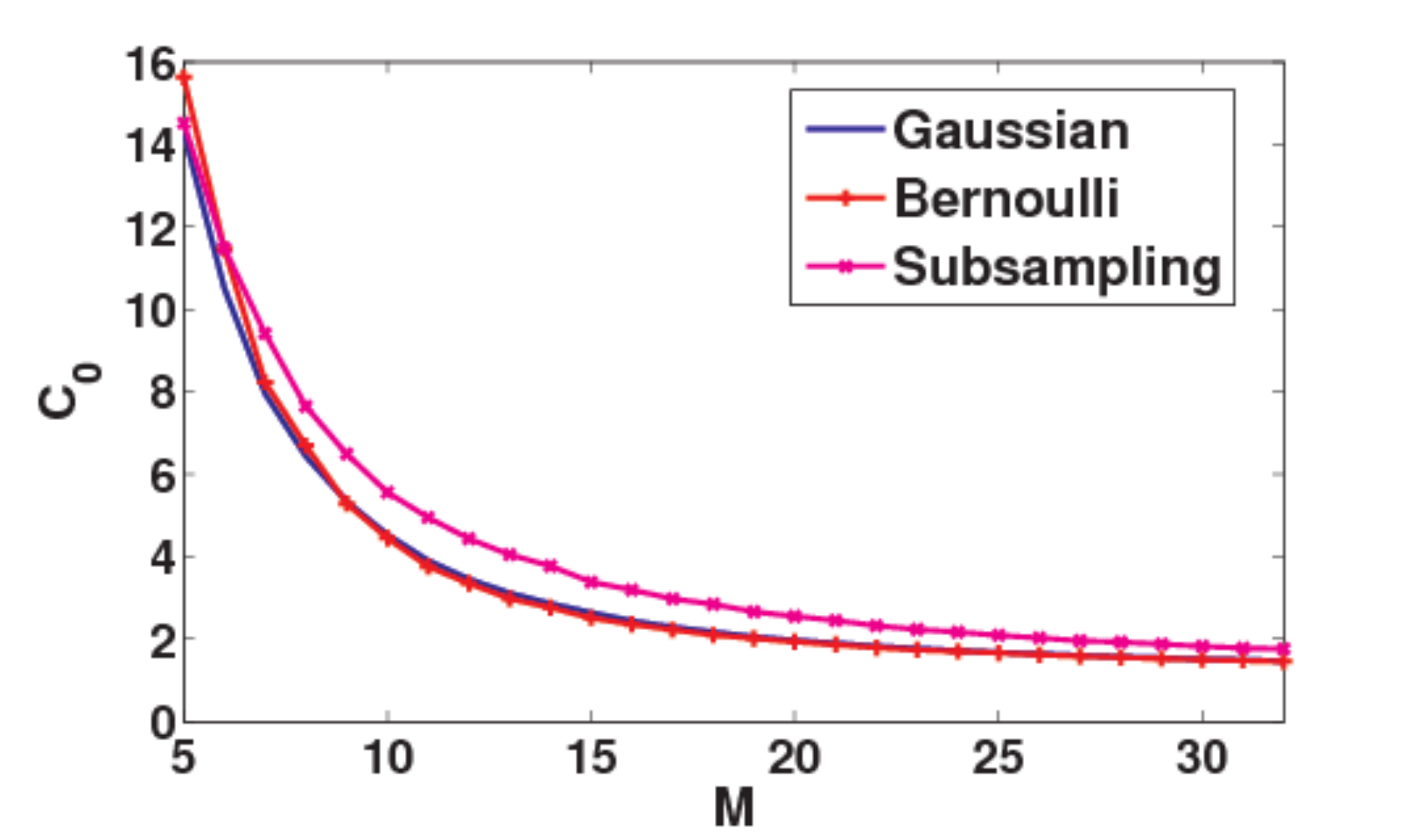}  &
\includegraphics[width=4cm]{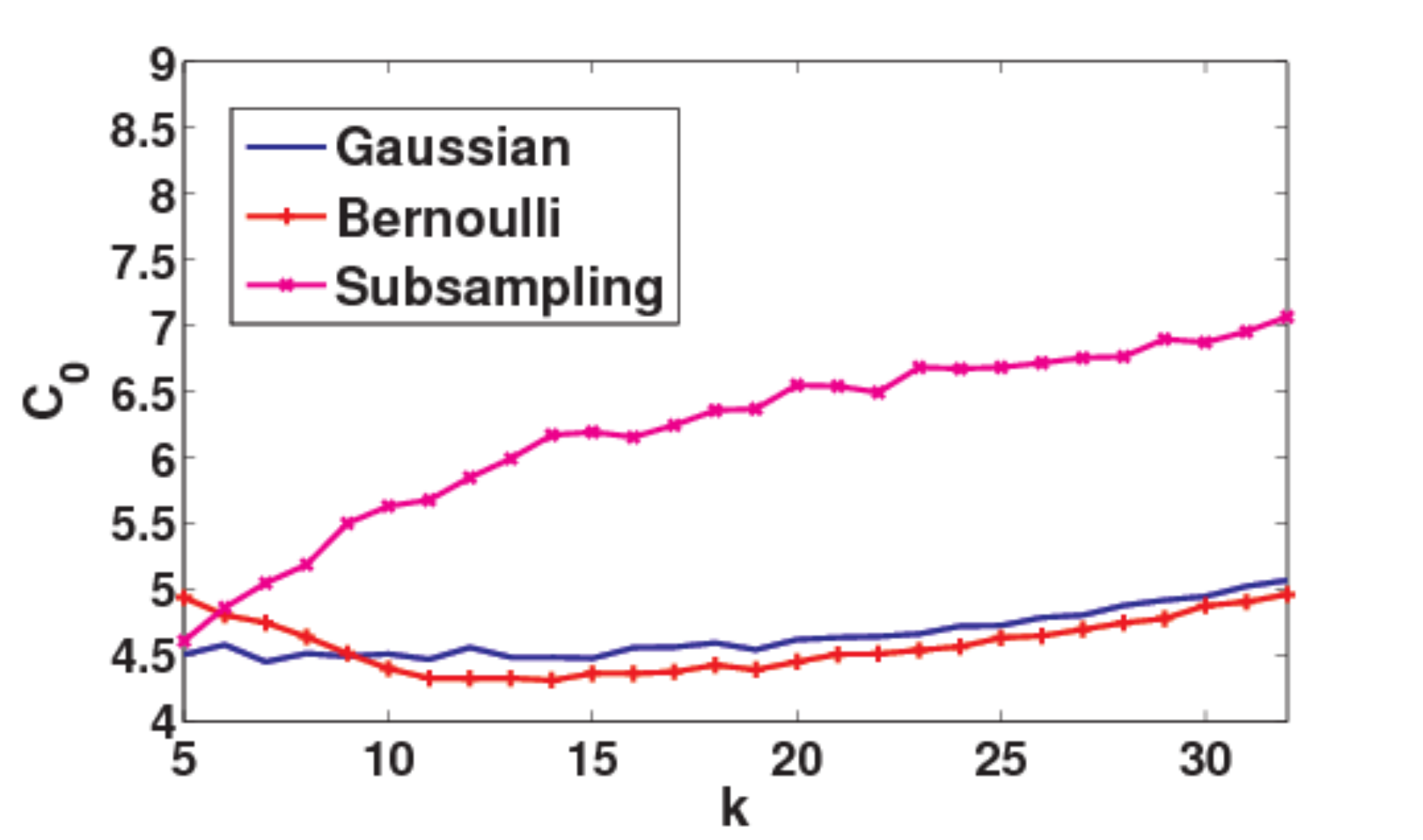} \vspace{-1ex} \\
\textbf{\small(a)} & \textbf{\small(b)} \vspace{-1ex} \\
\end{tabular}
\end{center}
\vspace{-3ex}
\caption{\small The MSE null space property constant $C_0$~\eqref{eqn:null:space:equality} of Gaussian, Bernoulli, and random subsampling matrices, as a function of $M$ (left), and of $k$ with $M=k$ (right).} \label{fig:RIP:expect}
\vspace{-4ex}
\end{figure}

Figure~\ref{fig:RIP:expect} (a) plots $C_0 = 1 + {b_K}/{a_K}$, obtained by simulating~\eqref{eqn:RIP:expt}, with $k=10$, for different values of $M$. When the number $M$ of SCS measurements increases, the reconstruction error of SCS decreases, resulting in a smaller ratio over the best-$k$ term approximation error with a fixed $k$. Gaussian and Bernoulli matrices lead to similar $C_0$ values, slightly smaller than that of random subsampling matrices. Figure~\ref{fig:RIP:expect} (b) plots $C_0$, as a function of $k$, with $M=k$. Gaussian and Bernoulli matrices lead to similar $C_0 \approx 4.5$ that varies little with $k$. For random subsampling matrices $C_0$ slowly increases, almost linearly, and is equal to $5.5$ for a typical value $k=10$, about 20\% larger than that of Gaussian and Bernoulli matrices.

From Corollary~\ref{corollary:nullspace2optimality:gaussian} and Theorem~\ref{theo:RIP:expect:null:space}, we obtain the next concluding result: 
\vspace{-4ex}
\begin{theorem}
\label{theo:SCS:with:RIP:expect}
Assume $\bx \sim \mathcal{N}(\bzero, \bS)$. Let $\Phi$ be an $M \times N$ sensing matrix and $\Delta$ the optimal and linear decoder~\eqref{eqn:MAP}. Then $\Phi$ satisfies the MSE instance optimality of order $k$~\eqref{eqn:instance:optimality:MSE} with constant $C_0 = 4(1 + {b_K}/{a_K})$, $a_K$ and $b_K$ given in~\eqref{eqn:RIP:expt}.\vspace{-1ex}
\end{theorem}

 \vspace{-4ex} 
\section{SCS with GMM \\ -- Experiments on Real Image Data}
\label{sec:numerics}
 \vspace{0ex} 
This section applies SCS with GMM in real image sensing and compares it with conventional CS based on sparse models. 

Following~\cite{yu2010PLE}, an image is decomposed into $\sqrt{N} \times \sqrt{N} = 8 \times 8$ patches $\{\bx_i\}_{1 \leq i \leq I}$ assumed to follow a GMM: There exist $K$ Gaussian distributions $\{\mathcal{N} (\mu_k, \Sigma_k)\}_{1 \leq k \leq K}$, and each image patch $\bx_i$ is independently drawn from one of these Gaussians with an unknown index $k$. SCS samples each patch $\by_i = \Phi_i \bx_i$, with a possibly different $\Phi_i$ for each $\bx_i$. The decoding scheme (i) estimates $\{(\mu_k, \Sigma_k)\}_{1 \leq k \leq K}$, and (ii) identifies the Gaussian distribution $k_i$ that generates the $i$-th patch, and reconstructs $\bx_i$ with the optimal and linear decoder~\eqref{eqn:MAP} associated to the appropriate Gaussian, $\forall i$. The resulting piecewise linear decoder is implemented with a computationally efficient MAP-EM algorithm alternating  between (i) and (ii)~\cite{yu2010PLE}. The algorithm has fast convergence and the MAP probability of the measured data increases as the algorithm iterates~\cite{yu2010PLE} (refer to~\cite{yu2010PLE} Sec. 2 for more details). 

The dictionary for conventional CS is learned with K-SVD~\cite{aharon2006k} from 720,000 image patches, extracted from the entire standard Berkeley segmentation database containing 300 natural images~\cite{MartinFTM01}. The decoder is calculated with an $l_1$ minimization. 

\begin{figure}[htbp]
\vspace{-2.5ex}
\begin{center}
\begin{tabular}{cc}
\hspace{-3ex}
\includegraphics[width=5cm]{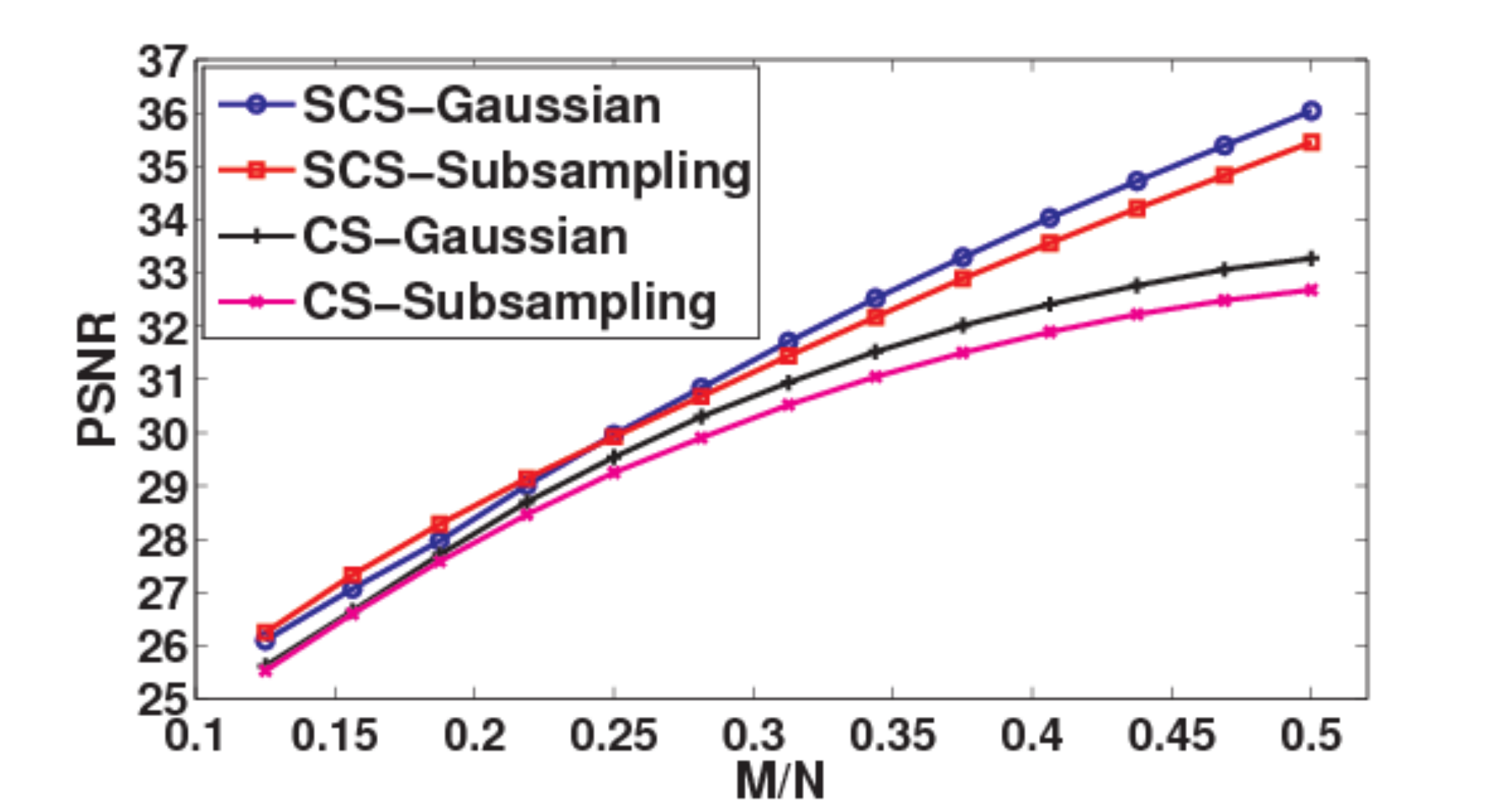}  \hspace{-5ex} &
\includegraphics[width=5cm]{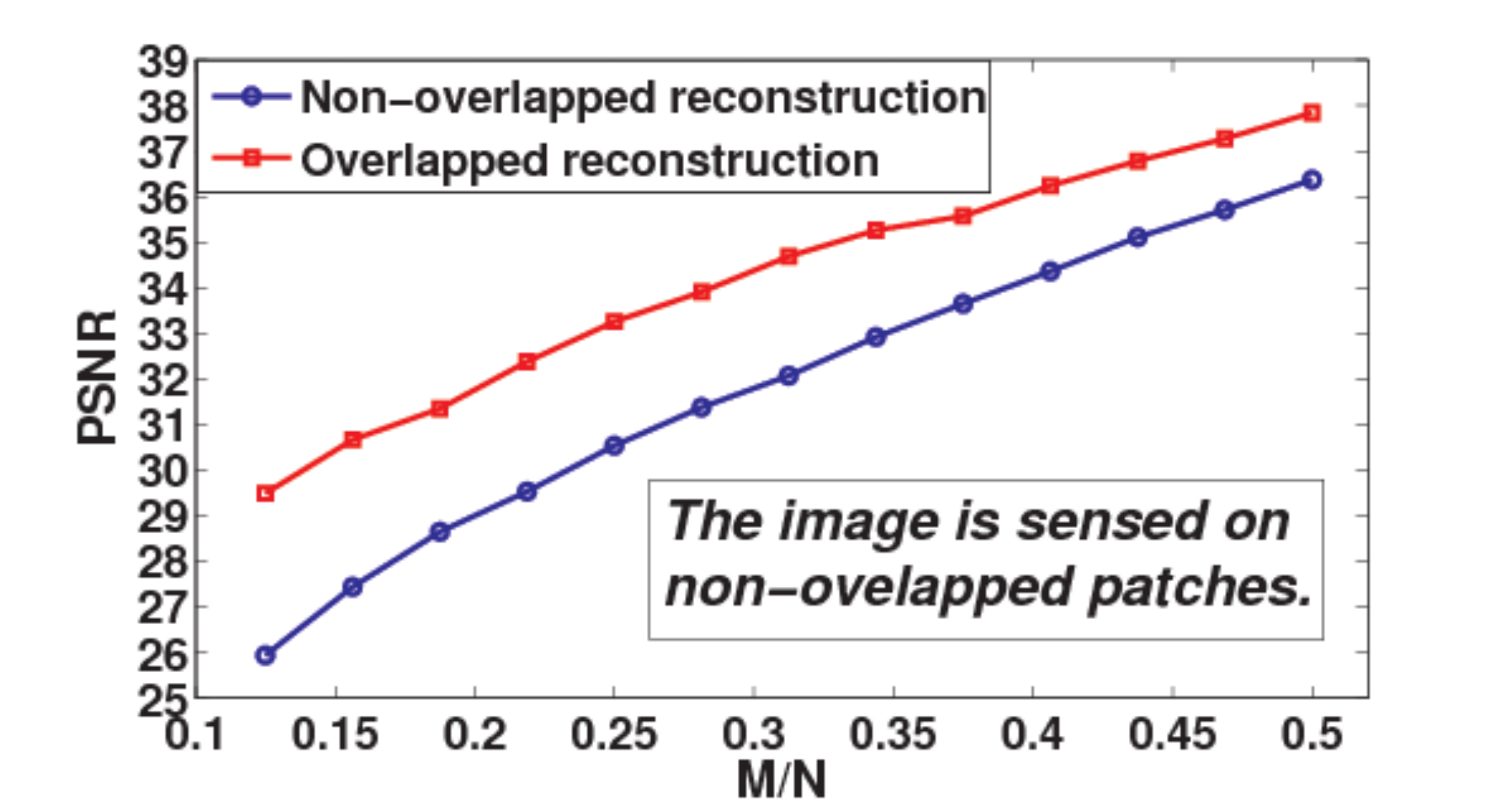} \vspace{-1ex} \\
\textbf{\small(a)} & \textbf{\small(b)} \vspace{-1ex} \\
\end{tabular}
\end{center}
\vspace{-3ex}
\caption{\small (a). Patch PSNR (dB) vs sampling rate for SCS and CS using Gaussian and random subsampling sensing matrices. (b) Image PSNR (dB) vs sampling rate, for SCS using Gaussian sensing matrices with non-overlapping reconstruction, and subsampling random matrices with overlapping reconstruction.} \label{fig:experiments}
\vspace{-2ex}
\end{figure}

\begin{figure}[htbp]
\vspace{-1ex}
\begin{center}
\begin{tabular}{cccc}
\hspace{-3.3ex}\includegraphics[width=2.5cm]{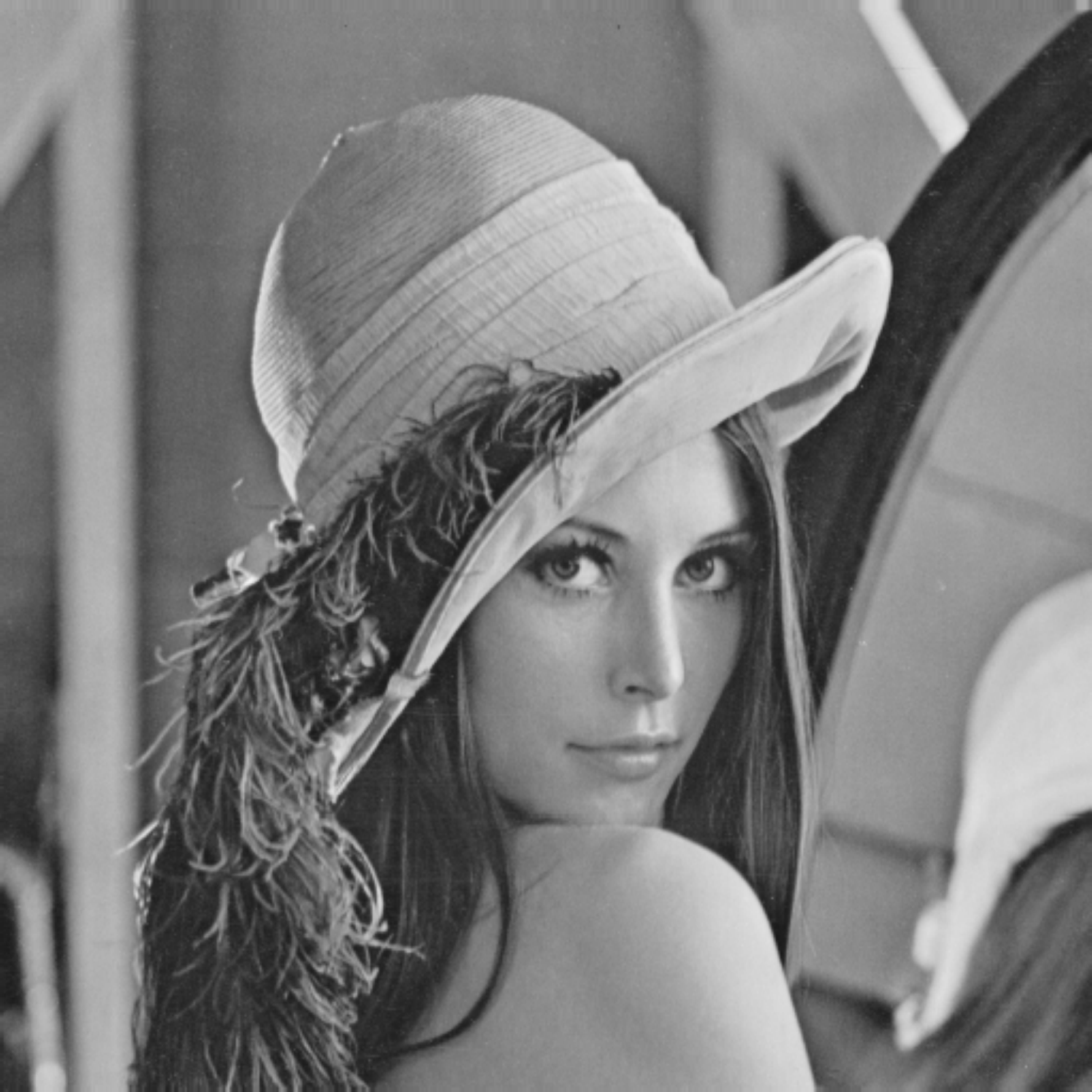}  &
\hspace{-2.3ex}\includegraphics[width=2.5cm]{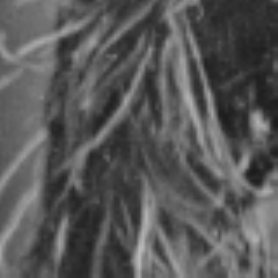}  &
\hspace{-2.3ex}\includegraphics[width=2.5cm]{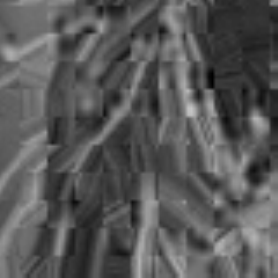}  &
\hspace{-2.3ex}\includegraphics[width=2.5cm]{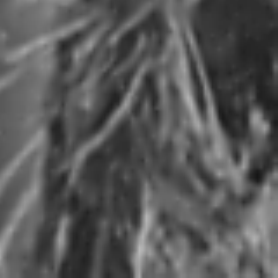}  
\vspace{-0.5ex} \\
\hspace{-3.3ex}{\scriptsize\textbf{Lena}} &\hspace{-2.3ex}{\scriptsize \textbf{Zoom (original)}}
&\hspace{-2.3ex} {\scriptsize \textbf{No.-ovl. rec. 30.54 dB}}&\hspace{-2.3ex} {\scriptsize \textbf{Ovl. rec. 33.26 dB}}
 \vspace{-1ex} \\
\end{tabular}
\end{center}
\vspace{-2ex}
\caption{\small From left to right. Lena, a zoomed crop, reconstructed image by SCS using Gaussian sensing matrices and non-overlapping reconstruction, and by SCS using subsampling random matrices and overlapping reconstruction. The image is sensed on \textit{non-overlapped} patches at a sampling rate of $M/N=0.25$.} \label{fig:Lena}
\vspace{-2ex}
\end{figure}

Figure~\ref{fig:experiments} (a) shows the sensing performance on about 260,000 (sliding) patches, regarded as signals, extracted from the standard image Lena, as shown in Figure~\ref{fig:Lena}. The PSNRs generated by SCS and CS using Gaussian and random subsampling sensing matrices, one independent realization for each patch, are plotted as a function of the sampling rate $M/N$. At the same sampling rate, SCS outperforms SC. The gain increases from about 0.5 dB at very low sampling rates ($M/N \approx 0.1$), learning GMM from the poor-quality measured data being more challenging,  to more than 3.5 dB at high sampling rates ($M/N \approx 0.5$). (SC using an ``oracle'' dictionary learned from the full Lena itself, undoable in practice, improves its performance from 0.2 dB at low sampling rates to 1.3 dB at high sample rates, still lower than SCS.) For both SCS and CS, Gaussian and random subsampling matrices lead to similar PSNRs at low sampling rates ($M/N<0.25$), and at higher sampling rates the former gains about 0.5 dB. Recall that SCS is not just more accurate and significantly faster, but also only uses the compressed image, while CS uses a pre-learned dictionary from a large database.

Figure~\ref{fig:experiments} (b) shows the sensing performance on the whole Lena image. The sensing is performed on \textit{non-overlapped} patches. Random subsampling matrices, which are diagonal operators (one non-zero entry per row), have the advantage 
of being able to lead to reconstruction on overlapped patches, averaging the overlapped reconstructed patches further improving the image estimation. The PSNRs generated by SCS using random subsampling matrices and overlapped reconstruction are plotted, in comparison with those obtained using Gaussian sensing matrices and non-overlapped reconstruction. The former improves from about 3.5 to 1.5 dB, at a cost of $N=64$ times computation. As illustrated in Figure~\ref{fig:Lena}, the overlapped reconstruction removes the block artifacts and considerably improves the reconstructed image. 

\noindent {\it \small \textbf{Acknowledgements:} Work partially supported by NSF, ONR, NGA, ARO, and NSSEFF. The authors thank very much St\'ephane Mallat for co-developing the GMM framework for solving inverse problems.}

\vspace{-2ex}
{\small
\bibliographystyle{plain}
\bibliography{biblio}}

\end{document}